\documentclass[12pt,openany,a4paper]{article}

	\usepackage[utf8]{inputenc} 
	\usepackage[english]{babel} 
	\usepackage[final,activate=true]{microtype} 

	\usepackage{libertine} 
	\usepackage{libertinust1math} 

	\usepackage{MnSymbol} 

	\usepackage[style=LingStyle,
		backend=bibtex,
		useprefix=true,
		sortcites=true,
		natbib=true,
		url=false,
		doi=false,
		minnames=1,
		maxnames=3,
		hyperref,
		dashed=false,
		]{biblatex}  

	\bibliography{references.bib} 

	\usepackage[style=english]{csquotes} 
	\usepackage{url}
	\usepackage[dvipsnames]{xcolor} 
	\definecolor{ColorOne}{named}{MidnightBlue} 
	\usepackage{hyperref} 
	\hypersetup{breaklinks,
		colorlinks,
		linkcolor=ColorOne,
		urlcolor=ColorOne, 
		citecolor=ColorOne,
		filecolor=ColorOne,
		plainpages=false,
		}	

	\usepackage{tikz} 
		\usetikzlibrary{arrows}

	\usepackage{tikz-qtree} 

	\usepackage{graphicx} 
		\graphicspath{{Grafiken/}} 

	\usepackage{linguex} 

\usepackage{todonotes}
\usepackage{algorithm}
\usepackage{algpseudocode}
\usepackage[capitalize]{cleveref}
\usepackage{subfig}
\usepackage[acronym, shortcuts, nohypertypes={acronym}]{glossaries}
\newacronym{AI}{AI}{artificial intelligence}
\newacronym{ASR}{ASR}{automatic speech recognition}
\newacronym{DDPM}{DDPM}{denoising diffusion probabilistic model}
\newacronym{DL}{DL}{deep learning}
\newacronym{DNN}{DNN}{deep neural network}
\newacronym{ESS}{ESS}{\emph{expressive speech synthesis}}
\newacronym{FM}{FM}{foundation model}
\newacronym{GenAI}{GenAI}{generative artificial intelligence}
\newacronym{HMM}{HMM}{hidden Markov model}
\newacronym{LAM}{LAM}{large audio model}
\newacronym{LLM}{LLM}{large language model}
\newacronym{MCMC}{MCMC}{Monte Carlo Markov Chain}
\newacronym{MFM}{MFM}{multimodal foundation model}
\newacronym{ML}{ML}{machine learning}
\newacronym{RL}{RL}{reinforcement learning}
\newacronym{RLHF}{RLHF}{reinforcement learning from human feedback}
\newacronym{SER}{SER}{speech emotion recognition}
\newacronym{SGM}{SGM}{statistical generative model}
\newacronym{SPSS}{SPSS}{statistical parametric speech synthesis}
\newacronym{TTS}{TTS}{text-to-speech synthesis}
\newacronym{VAE}{VAE}{variational autoencoder}

\newcommand{\cf}{{cf.\ }}

\begin{document}

\title{Expressivity and Speech Synthesis}
\author{Andreas Triantafyllopoulos, Björn Schuller}
\maketitle


\section{Introduction}
Talking machines have long captured human imagination.
Already in Homer's Iliad (8th century BC) we read of Hephaestus' ``golden maids''~\citep[Hom. Il. 18.388]{Bassett25-HTI} (emphasis ours):
\begin{quote}
    Waiting-women hurried along to help their master [Hephaestus].
    They were made of gold, but looked like real girls and \emph{could not only speak} and use their limbs but were also endowed with intelligence and had learned their skills from the immortal gods.
\end{quote}
That they could speak was seen as a prerequisite of intelligence (and being worthy of serving in a god's retinue).
The first documented efforts to endow machines with the capacity to speak date back to late 17th century AC, with C. G. Kratzenstein's vowel resonators~\citep{Ohala11-CGK}.
Arguably though, the quest for human-like speech synthesis began in earnest with the advent of computers in the 1950-1960s, with 1961 seeing the first digital vocoder implemented on an IBM 7090~\citep{Kelly61-AAT}.
Ever since, speech synthesis research has predominantly focused on producing utterances that accurately convey the target text; in a nutshell, its end-goal is a system that accepts as input a text sequence and produces an audio signal such that, when humans are exposed to it, they will decode it into the identical text sequence.

Despite the fact that even early systems were able to produce speech that was, to a certain extent, understandable, they were severely lacking in \emph{naturalness}.
This led to a call for improving the \emph{expressivity} of synthesised speech; in this context, expressivity was defined as the mimicking of prosodic, rhythmic, and other paralinguistic patterns exhibited by humans.
In a sense, this imitation was initially \emph{directionless}, i.\,e., researchers aimed to copy or approximate human prosody and rhythm without any direct communicative intent.
This approach has left its mark even on contemporary research and how it approaches expressivity, as we discuss in \cref{ssec:control}.

However, there has been a long, related tradition of research on the additional communicative and informative functions of speech beyond its linguistic aspects~\citep{Scherer86-VAE, Schuller14-CPE}.
These additional phenomena can be collectively referred to as the \emph{paralinguistic} component of speech.
As discussed in \citet{Schuller14-CPE}, paralinguistics can be defined to subsume \emph{extralinguistics} to cover a wide gamut of phenomena, from informative functions regarding nearly immutable speaker characteristics, like age or gender, to communicative functions regarding short-term states such as (political) stances and emotions.
From this perspective, \ac{ESS} can then be seen as the \emph{purposeful} attempt to imitate specific \emph{states} and \emph{traits} through the manipulation of acoustic and prosodic variables in the synthesised utterance.
This is the main topic of this chapter.

In terms of technical advances, the field has come a long way since the primitive vocoders of the early computer era.
Starting with model- and `rule'-based approaches~\citep{Kelly61-AAT, Coker76-MAD}, quickly moving to data-driven concatenative synthesis~\citep{Allen79-MIT, Klatt87-TTS, Moulines90-PSW, Khan16-CSS}, and then later to statistical models~\citep{Tokuda00-SPG, Zen09-SPSS}, \ac{TTS} has progressed in leaps-and-bounds in recent years with the advent of \ac{DL}~\citep{Tan21-NSS}.
\Ac{ESS} followed a parallel developmental path to standard speech synthesis, with Cahn's Affect Editor~\citep{Cahn89-AE, Cahn90-AE} and Murray's HAMLET~\citep{Murray89-HAMLET, Murray93-TTS} representing the earliest methods relying on rules, while later approaches transitioned to concatenative~\citep{Schroeder01-ESS, Iida03-ACB}, parametric~\citep{Tachibana04-HMM, Tao06-PCF}, and, finally, \ac{DL}-based~\citep{Triantafyllopoulos23-AOO} \ac{ESS}.
With each change in technology came associated gains in fidelity and naturalness~\citep{Triantafyllopoulos23-AOO}.

This trend is exemplified by the recent wave of advances in the broader \ac{GenAI} field~\citep{Fui23-GAI}.
Progress in probabilistic generation, currently spearheaded by ``diffusion models''~\citep{Yang23-DMA}, have brought \ac{GenAI} in the epicentre of attention for various stakeholders -- societal, commercial, and, increasingly, regulatory (see e.\,g., the recent EU AI Act~\citep{AIAct}).
Text generation has been the most prominent example of that new era, with \acp{LLM} like ChatGPT~\citep{Achiam23-GPT}, Llama~\citep{Touvron23-LOF}, or Claude spearheading recent innovations.
Mirroring that success, the quest for \ac{ESS} breakthroughs is being taken on by an increasing number of research groups and companies, and has become a staple of speech technology conferences (INTERSPEECH, ICASSP, SLT, etc.).

Expectedly, synthesis quality and controllability are improving at an accelerating rate~\citep{Triantafyllopoulos23-AOO}.
Moreover, as a result of increased commercial interest, \ac{ESS} systems of unprecedented capabilities are being constantly released to the public, in off-the-shelf, easy-to-use toolkits that can be co-opted by a wider and wider cohort of lay users for their own purposes.
On top of that, \emph{foundation models} have recently surfaced as a key differentiator in \ac{GenAI} and beyond~\citep{Bommasani21-OTO} and are beginning to impact \ac{ESS} as well~\citep{Yang23-UAA}.
This means that we will soon be living in a ``metaverse''~\citep{Mystakidis22-MET} populated with expressive \ac{AI} agents whose voices are indistinguishable to humans, and whose capabilities may vastly exceed (or enhance) the voices of average people.
Accordingly, this increases the probability that bad actors, or even well-intentioned users, misuse the technology -- a problem encompassed in the broader \ac{AI} ``alignment'' conversation~\citep{Gabrial20-AIV}.

Beyond that, the present situation begs the question: \emph{What else remains to be done?}
As we argue, contemporary research is largely geared towards ``expressive primitives'' -- states and traits which are straightforward to depict and can be simulated within a singular utterance -- and which we call \emph{Stage I} \ac{ESS} research.
Typical examples include the synthesis of ``emotional voices'': this results in speech which will be perceived as conveying a particular emotion (e.\,g., happiness).
However, a major promise of \ac{ESS} systems lies in facilitating a conversational interface between humans and \ac{AI} agents.
In fact, given the rise of modern text-based conversational agents (i.\,e., `chatbots') like ChatGPT~\citep{Achiam23-GPT}, we expect \ac{ESS} systems to become embedded in voice-driven conversation applications, where emulating an emotional state goes beyond portraying that emotion within a particular utterance.
In other words, \emph{appropriateness} becomes an essential aspect -- what to say, when, and how.
What is more, there are expressive states which cannot be distilled to a single component, such as political stances, moods, or dispositions~\citep{Schuller14-CPE}.
Given the anticipated mastering of synthesising unitary utterances, we expect an increased focus on synthesising more nuanced, longer-term states and traits of expressive agents, as well as adapting to the context of (real-time) conversations with different individuals.
This we call \emph{Stage II} research, and it is still in its nascent stages.

This chapter aims to chart this emerging landscape of expressivity in the era of \ac{GenAI}.
With that in mind, our goal is \emph{not} to give a technical survey of state-of-the-art systems, as there exist a plethora of older and more recent surveys that sketch out the inner workings of \ac{TTS} and \ac{ESS} approaches over the years; we recommend \citet{Schroeder01-ESS, Zen09-SPSS, Khan16-CSS, Tan21-NSS, Triantafyllopoulos23-AOO, Yang23-DMA, Barakat24-DLE} as good starting points.
Therefore, we intentionally place limited emphasis on the technical implementations of existing \ac{ESS} systems.
Instead, we explore deeper questions that are highly pertinent for the present and future of the field.
Among others, we discuss:
\begin{itemize}
    \item What are the states and traits that we can expect \ac{ESS} systems to cover (\cf \cref{sec:taxonomy})?
    \item How can we move from classic, simple expressive `primitives' (\cf \cref{sec:primitives}) to more complex behaviours? 
    How can we jointly synthesise multiple -- perhaps contradictory -- states (\cf \cref{sec:complex})?
    \item How can we move away from a `one-size-fits-all' approach and towards a more personalised approach to synthesis (\cf \cref{ssec:personalisation})?
    \item What role do foundation models play (\cf \cref{sec:foundation})?
    \item Crucially, what happens to the world once we achieve our wildest dreams regarding the capabilities of \ac{ESS} models?
\end{itemize}
We hope that our discussion of these questions will help shape future research and provide a template and roadmap for the next generation of \ac{ESS} systems.
Importantly, our discussion is grounded in the applications that \ac{ESS} enables, as these dictate its \emph{ecology} and thus the \emph{affordances} that it may develop, and so set, in turn, the framework for current and future research efforts.

The remainder of our chapter is organised as follows.
We first introduce a taxonomy of states and traits which can be expressed in speech and which, accordingly, \ac{ESS} aims to simulate, followed by a discussion of the applications which it facilitates.
Next, we describe the technical underpinnings of traditional and contemporary systems.
Following that, we discuss our notion of a \emph{Stage II} system and review how foundation models are utilised in this field.
Our final section outlines the ethical considerations entailed by a rapidly advancing technology.

\section{Expressivity in speech synthesis}
In this section, we first present the states and traits which can be synthesised with \ac{ESS}, and then the application domains in which \ac{ESS} has found, or can be expected to find, widespread adoption.

\subsection{A taxonomy of expressive states and traits}
\label{sec:taxonomy}

\begin{figure}[t]
    \centering
    \includegraphics[width=\textwidth]{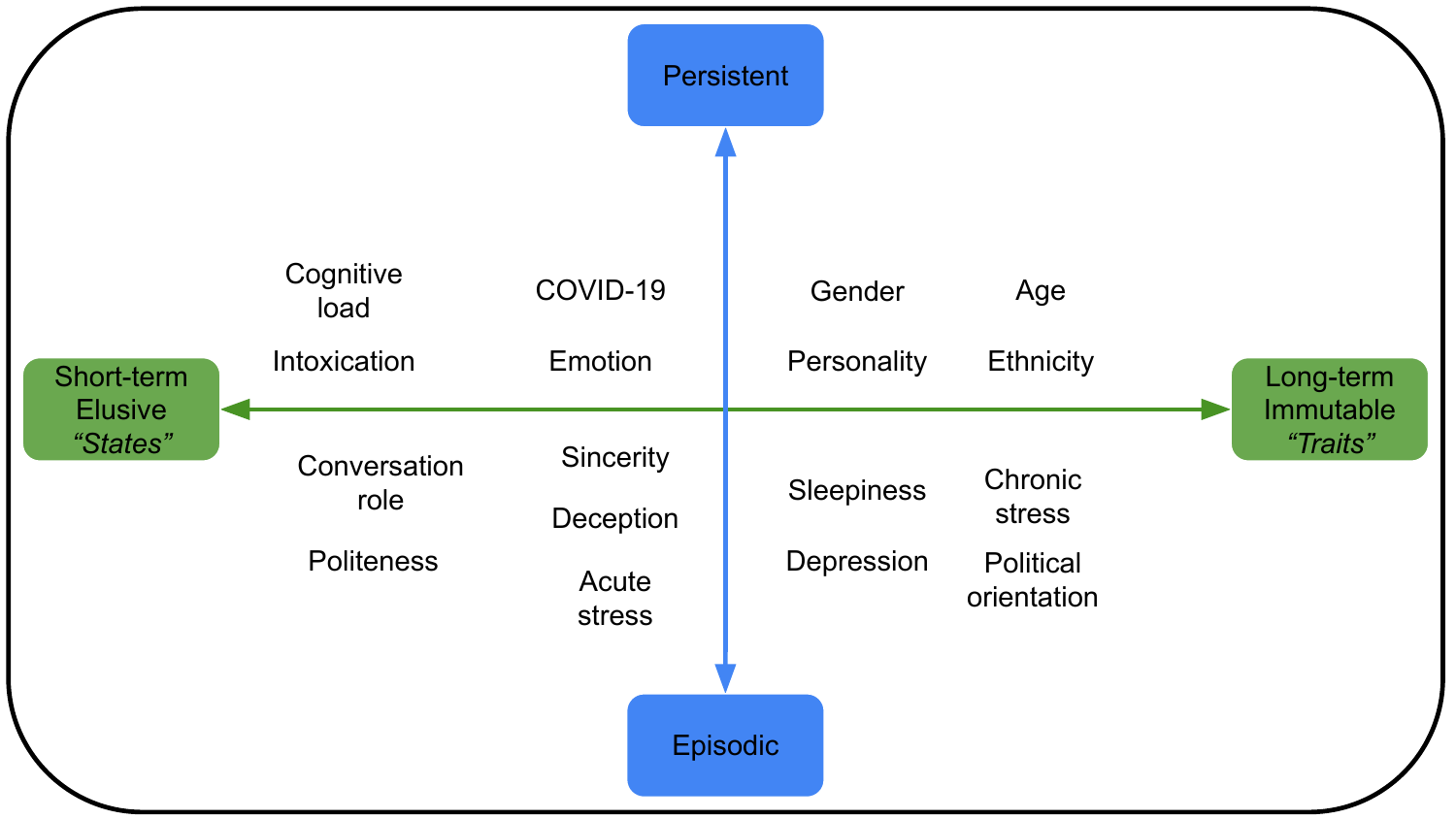}
    \caption{
    A non-exhaustive taxonomy of states and traits that \textbf{humans} express through their speech largely informed by previous work on recognising them (e.\,g., see \url{http://www.compare.openaudio.eu/tasks/} as well as \citet{Schuller14-CPE}).
    While these states might \underline{not} all be relevant for \ac{ESS} systems, they illustrate the plethora of styles that can be synthesised.
    Further, they help us distinguish between two crucial components -- how long each style lasts, and how persistent its appearance is.
    }
    \label{fig:taxonomy}
\end{figure}
We begin with a basic assumption: that the things which can be recognised are those that will also be (eventually) synthesised.
If humans, and by extension \ac{AI} algorithms, can recognise particular affective states in speech, then there is nothing, in principle, preventing \ac{GenAI} algorithms from simulating those states.
This definition allows us to sidestep what \emph{has} been done by the community (due to various restrictions including the lack of available data) vs what \emph{can} be done (based on evidence that humans can portray particular states).

\cref{fig:taxonomy} presents a non-exhaustive portrayal of expressive styles that can be recognised in humans -- this serves as inspiration for all that can be potentially synthesised for computers.
It is motivated by similar taxonomies outlined in \citet{Scherer03-VCO} and \citet{Schuller14-CPE}.
Importantly, we distinguish between two particular axes\footnote{Note that there are other axes in which these styles differentiate themselves, such as whether they are event-focused or elicit an appraisal response~\citep{Scherer03-VCO}.}:
how long the underlying affective state lasts and how persistent it is in its appearance.

The first axis allows us to decompose affective behaviour into \emph{states} and \emph{traits}~\citep{Schuller14-CPE}.
At the extreme, states are short-lived, transient conditions; these include concepts like emotion or interest.
Traits, on the other hand, are more long-term, less mutable attributes, such as gender or personality.
Naturally, this taxonomy is not a black-and-white dichotomy, but a colourful spectrum: in-between the two extremes exists a variety of conditions that vary with respect to their duration, intensity, and (expected) rate of change.

The second axis instead differentiates how prevalent a state or a trait is in a person's speech.
Some behaviours are \emph{persistent}; at the extreme, they are ever-present, and ubiquitously colour almost every single utterance.
On the other end are \emph{episodic} behaviours; those are fleeting, appearing only momentarily and within the confines of a single expression.
Crucially, this second distinction is independent of whether the behaviour is a state or a trait.
Some traits, like gender or age, are both long-term and persistent (they are almost always to be detected in a speaker's voice); others, like political orientation or depression are \emph{episodic}\footnote{A depressed individual will not be sad all the time and even the most ardent political activist will occasionally take a break from street protests.}.
Likewise, even though states themselves are fleeting, some, like politeness or sincerity, are to be found in a small set of utterances, while others, like emotion or respiratory diseases are constantly present -- so long as they continue to be true for a speaker\footnote{The interaction of the two axes where a short-lived state appears intermittently results in \emph{transient} behaviours, or, equivalently, limits the amount of episodes to 1.}.
This temporality is important for understanding affective behaviour in humans, as, like \citet{Cowie03-DTE} argues:
\begin{quote}
    A real possibility is that there may be multiple scales at work even in the short term, with some signs building up over a period of seconds or minutes and others erupting brieﬂy but tellingly.
\end{quote}

This important aspect of timing also calls for different \ac{ESS} capabilities.
The first, simpler one, requires a mapping from one state to another; this is applied consistently to all utterances.
We call this \emph{Stage I} \ac{ESS}, and it is primarily suitable for synthesising persistent behaviours.
The second, more sophisticated one requires knowing which utterance to transform into a particular expression; often, the desired effect is achieved by combining multiple simple expressions from \emph{Stage I}.
We call this \emph{Stage II} \ac{ESS}, and it is geared towards long-term, episodic behaviours.

\emph{Stage II} is much more challenging than \emph{Stage I}.
Specifically, while the generation of a persistent expression \emph{in situ} is possible using a direct portrayal of it, utilising such a portrayal \emph{in context} to reflect an episodic expression is substantially more challenging~\citep{Clark19-WMA}.
Seen in this light, contemporary \ac{ESS} systems have mastered the synthesis of behavioural `primitives' -- affective states that can be portrayed fleetingly, oftentimes within a single utterance or even within a single word.
However, these primitives form the basic building blocks for an array of more complex states and traits that have thus far remained elusive, like personality, ideology, stance, friendship, or compassion.

\subsection{Application domains for ESS}
\label{sec:applications}
\begin{figure}[t]
    \centering
    \subfloat[Human-computer interaction]{\includegraphics[width=.45\textwidth]{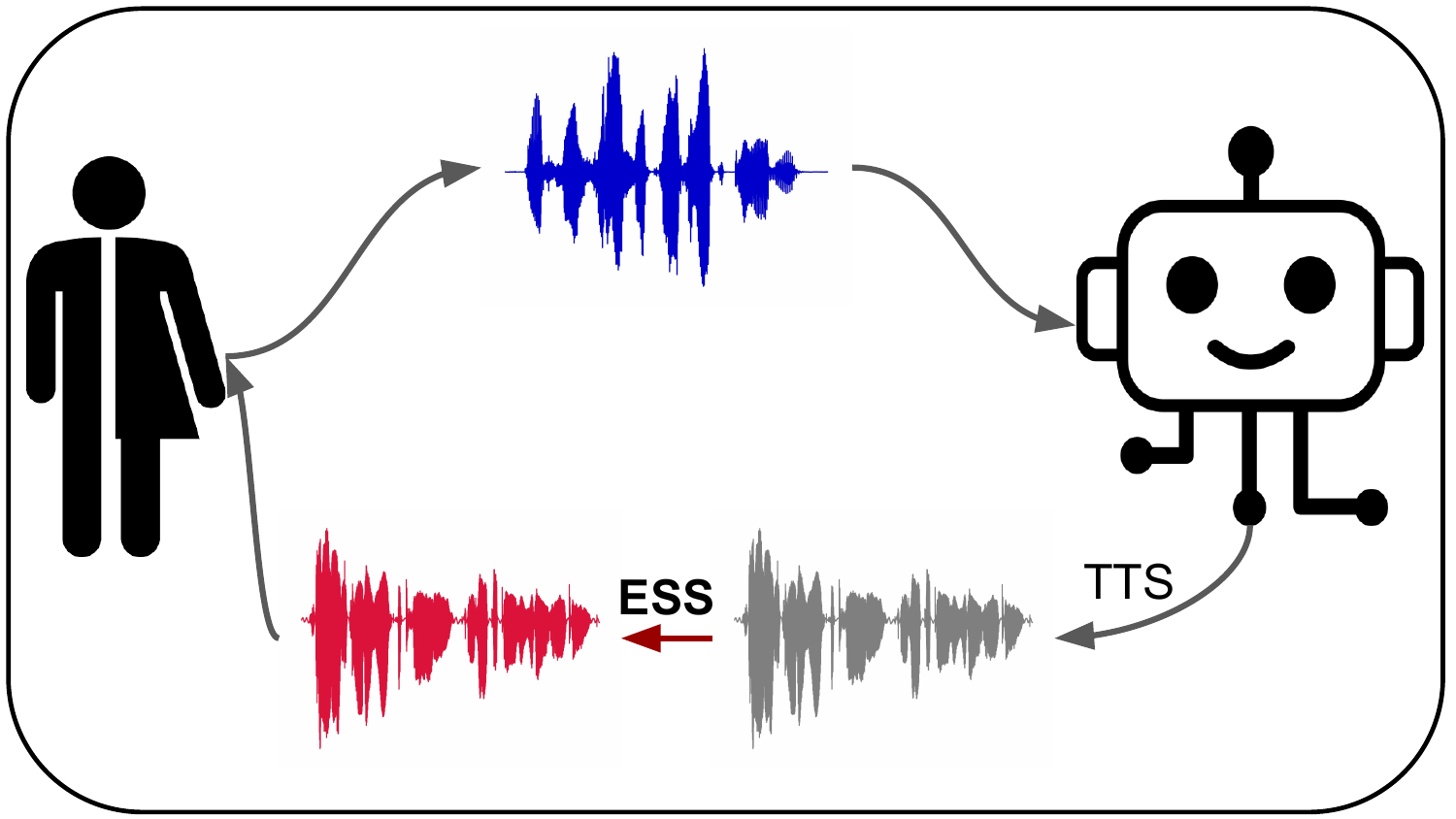}}~%
    \subfloat[Content creation]{\includegraphics[width=.45\textwidth]{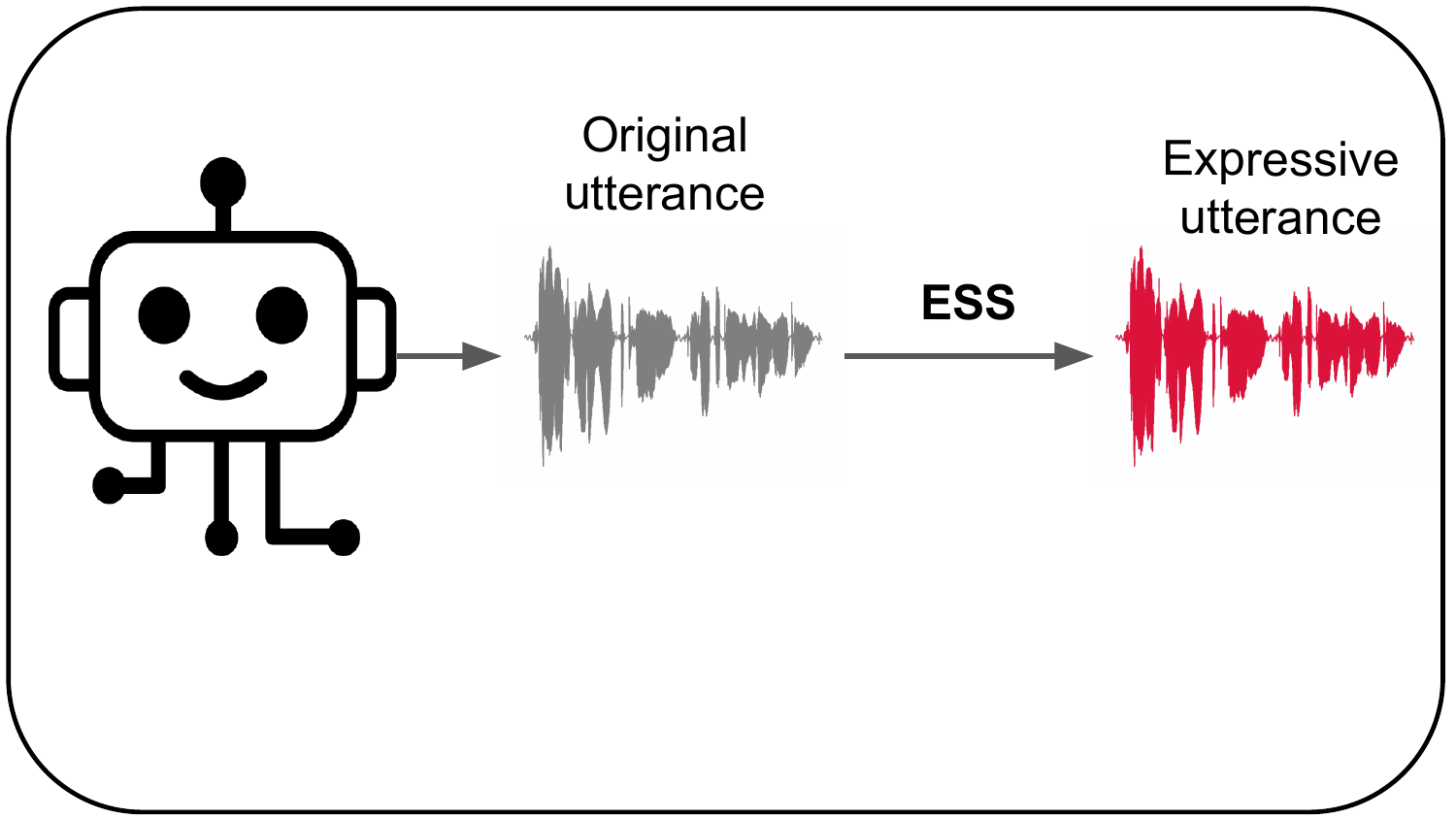}}
    
    \subfloat[Voice enhancement]{\includegraphics[width=.45\textwidth]{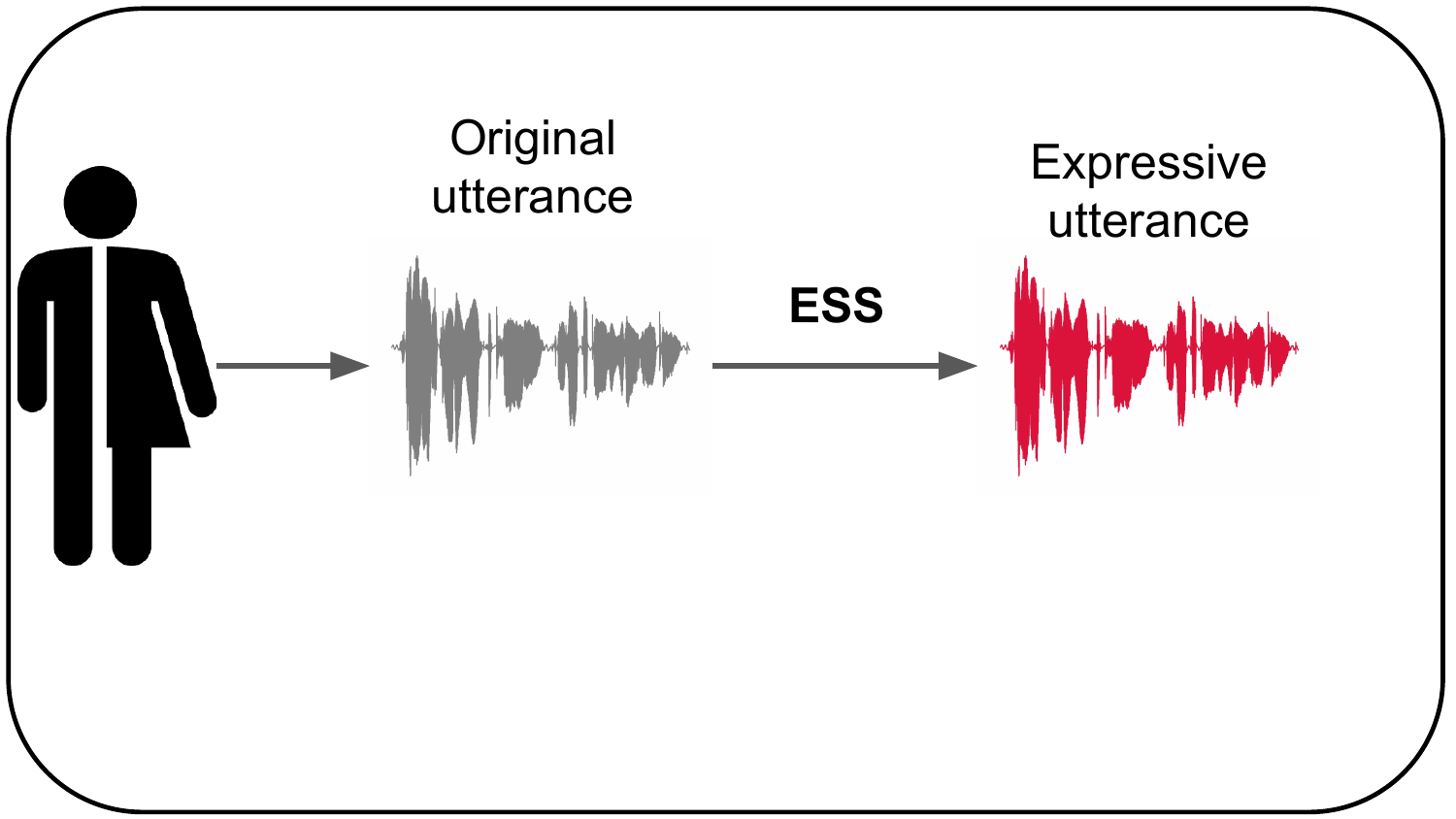}}~%
    \subfloat[Computer-computer interaction]{\includegraphics[width=.45\textwidth]{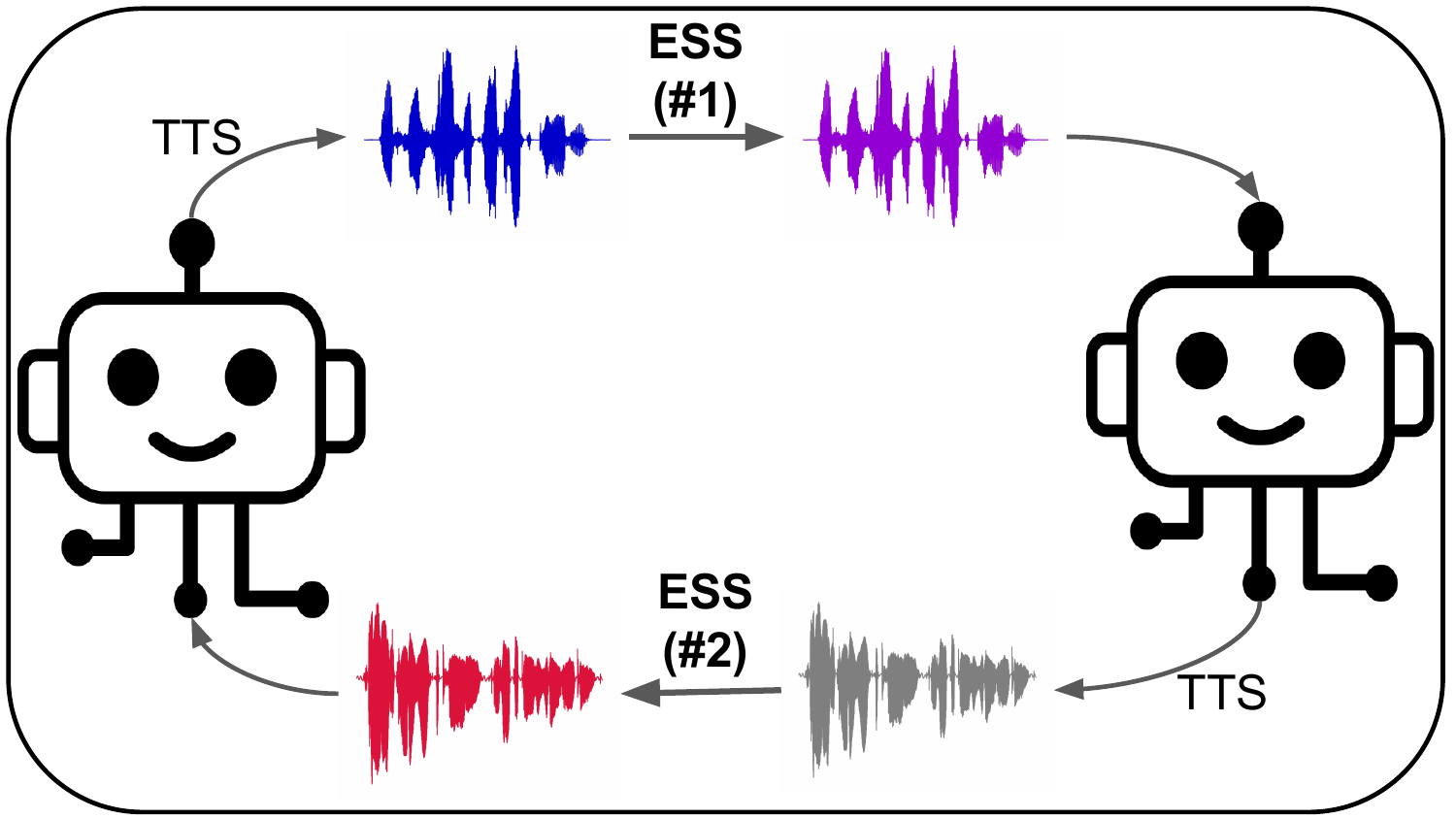}}
   \caption{
   Overview of different application domains that can benefit from expressive speech synthesis:
   Human-computer interactions entails the real-time communication between a human and an expressive chatbot;
   Content creation encompasses all possible forms of \emph{de novo} artificial content creation (e.\,g., video narration);
   Voice enhancement is targeted to the manipulation of a real human's voice;
   Finally, computer-computer interaction sketches a scenario where expressive chatbots communicate with one another in-the-wild.
   }
   \label{fig:applications}
\end{figure}

Having reviewed the states and traits that \ac{ESS} can be expected to simulate, we now turn to its potential real-world applications.
We consider application domains where use of \ac{ESS} is already widespread, others, where text-based communication is a given reality, with speech being the natural extension, and, finally, ones which have not yet seen much use of \ac{ESS}, but are ripe for disruption.
Some of the fields we discuss below saw the widespread use of text-based technologies even before the introduction of \acp{LLM}.
The unprecedented capabilities offered by those have naturally disrupted previous standard processes and made their integration a very active area of ongoing research.
It is in this landscape that we discuss \ac{ESS} applications.

An overview of the four types of application domains that are highly relevant for \ac{ESS} research is given in \cref{fig:applications}.
\ac{ESS} systems are, so far, primarily used to facilitate human-computer interaction.
In that scenario, it is typical to combine \ac{ESS} with \ac{TTS}, and create entirely artificial voices.
However, \ac{ESS} can also be used to transform existing human voices~\citep{Stylianou09-VTA} -- an application which is similar from a technical perspective but has different societal implications -- or even generate entirely artificial content, but outside the context of a conversation (e.\,g., for marketing).
We discuss these three scenarios below.
Moreover, we briefly mention the exotic case of computer-computer interaction as an emerging new frontier for \ac{ESS} research.

\subsubsection{Human-computer interaction} 
First and foremost on the list of current and future applications are conversational agents.
Several companies employ text chatbots to offload some of their workload for customer support, service, or even sales~\citep{Lester04-CAG}, and this field is rapidly growing with the rise of \acp{LLM}.
Moreover, different institutions see the promise of conversational agents in improving their services, like seen in the healthcare domain~\citep{Laranjo18-CAI}.
Finally, intelligent assistants (e.\,g., Siri, Alexa, or Google Assistant), by now pervasive in various consumer gadgets (like smartphones or even smartwatches), are essentially more advanced chatbots, with most of those including \ac{TTS} in their workflow.

Further integrating \ac{ESS} capabilities to all these conversational agents is a straightforward extension of their present state~\citep{Asghar18-ANR, Hu22-TAE}.
We thus expect this application domain to be both a key driver and an early adopter for future advances.
In terms of expressivity, conversational agents may place an emphasis on interpersonal adaptation to the user, appearing helpful, empathetic, or show any other personality trait that is desirable to their creators.

\subsubsection{Content creation}
Besides interacting with humans, \ac{ESS} can be used to facilitate the \emph{de novo} creation of new content, especially when combined with the impressive capabilities of broader \ac{GenAI} models.
Marketing is a domain seeing increased use of \ac{GenAI} technologies~\citep{Kshetri23-GAI}, where employing \ac{ESS} to accompany automatically generated illustrations or videos is attracting community attention.
In the extreme, this can go as far as creating entirely virtual personas, or artificial \emph{influencers}, which populate digital spaces and promote marketing material in a more naturalistic way than a simple commercial can ever do~\citep{Sands22-FIU}.

On the darker side of speech science, \ac{ESS} can vastly expand the capacity of bad actors to spread misinformation.
This can be done as a straightforward case of ``marketing'', albeit for an evil cause, with \ac{ESS} being used for promotional material around fake news in the same way as a company may use it to promote its product.
For example, \emph{vishing} -- the use of voice calls for phishing -- is one area ripe for disruption from \ac{ESS} software~\citep{Krombholz15-ASE}.
In its present form, performed by humans, it is already a major societal and legal problem\footnote{See a recent report by the United States' Federal Bureaue of Investigation: \url{www.ic3.gov/Media/PDF/AnnualReport/2021_IC3Report.pdf}.}, and is bound to get worse as \ac{ESS} facilitates a scaling up of resources available to bad actors that engage in this practice.

\subsubsection{Voice enhancement}
\ac{ESS} can also be used to augment or enhance one's own voice to attain specific expressive attributes that it is lacking.
This can be done both short-term, e.\,g, when one sends a short voice message to their partner and wants to convey some additional affect they are not able to express at the moment, such as excitement for an upcoming dinner that they are not presently feeling due to fatigue, but also long-term, e.\,g., to manipulate one's entire persona for a social media profile.
For example, female politicians have sometimes undergone intentional training to change their manner of speaking, with Hilary Clinton and Margaret Thatcher purportedly switching to a more masculine voice~\citep{Cameron05-LGA, Jones16-TLA}.
In the future, this may be achieved by a simple application of voice conversion.

Beyond one's self, however, \ac{ESS} can be used to transform the voice of others -- usually for malicious purposes.
This is essentially a more subtle form of \emph{deep faking}~\citep{Chesney19-DFA}, where instead of using \ac{GenAI} to fabricate a non-existent statement from an individual, one may use voice transformation to distort the original message.
As a recent example, much debate revolved around the age of the United States president at the time of writing, Joe Biden.
A similar use of voice technology could make him appear older than he actually is, thus intensifying his opponents' accusations regarding his suitability as a candidate.
This more subtle form of manipulation is perhaps more subversive than outright fakes; while those can be vetted and refuted based on evidence and facts, minute changes to the voice of a speaker that cast them in a negative light will be much harder to identify.
Broadly, this subtle transformation of one's voice can be used to harm political or commercial opponents, or even entire social groups, and we expect it to become a major societal issue in the future.

\subsubsection{Computer-computer communication} 
Finally, we want to highlight that in a world were conversational agents and intelligent assistants are increasingly deployed to perform more complex tasks autonomously, such as booking appointments or handling business transactions, we can expect that these systems will also encounter artificial interlocutors throughout their lifecycles.
For example, one person's intelligent assistant might attempt to book an appointment over the phone with a company's artificial customer service agent.
In that case, both systems might presumably use \ac{ESS} to achieve the desirable outcome while remaining oblivious to the fact that they are communicating with another machine.
This raises interesting implications both on how these systems will react and the types of affordances they will need to develop for success (assuming they are to some extent learning autonomously).

\section{Stage I: Synthesising expressive primitives}
\label{sec:primitives}
In this section, we discuss the technical aspects behind generating \emph{short} utterances that convey one particular expressive state, beginning with a historical overview and continuing with a discussion of how the basic principles of generative models can be applied to the generation of speech and vocal bursts.
We provide more background on stochastic generative models in \cref{ssec:sgm} and extend this discussion with more technical details in \cref{ssec:sgm}.
We additionally provide an overview of how speech utterances and vocal bursts are synthesised.
The section finishes with a discussion of the controllability of \ac{ESS} models.

\subsection{A blitz history lesson}
\label{sec:history}
Expressivity was embedded in \ac{TTS} systems from their first incarnation.
The first vocoder by \citet{Kelly61-AAT} allowed for the manipulation of prosodic and timbre attributes that are related to affect~\citep{Scherer03-VCO, Schuller14-CPE}.
However, the first documented attempts to use this functionality came much later, with rule-based systems like HAMLET~\citep{Murray89-HAMLET, Murray93-TTS} and Affect Editor~\citep{Cahn89-AE, Cahn90-AE}.
These manipulated attributes which are known to correlate with affect, such as pitch or timing.
These parameters were later investigated in a data-driven fashion~\citep{Burkhardt00-VAC}, but, by and large, this first era of \ac{ESS} largely depended on rules and expert knowledge.

The next generation featured concatenative synthesis~\citep{Van97-PIS, Schroeder01-ESS, Black03-USE, Iida03-ACB}, which relied on selecting speech units uttered with the appropriate expressions from an existing corpus.
We note that at this point, the `sister' field of \ac{TTS} had already progressed to \ac{SPSS}, where trainable modules are learnt from data, and this was readily co-opted for \ac{ESS} too~\citep{Tachibana04-HMM, Tao06-PCF}.
Typically, these models were implemented with \acp{HMM} and were trained to map prosodic and spectral features from a `neutral' to an expressive state.
Most often, this necessitated a cascade pipeline, with the initial synthesis made with a standard \ac{TTS} tool and an extra conversion \ac{ESS} module applied on top.
Learning this transformation usually required \emph{parallel} data -- corpora containing audio pairs which only differed in the expressed state but everything else (speaker, text) remained the same.

With the advent of \acl{DL}, \acp{HMM} were substituted with \acp{DNN}~\citep{Triantafyllopoulos23-AOO}, but the key principles remained the same.
A mapping from some neutral state to an expressive one was learnt from data.
However, the capabilities of \acp{DNN} further allowed for a disentanglement between the different components of speech, thus no longer demanding parallel data.
This enabled a radical scaling-up of the available speech that could be used for training, an increase which went hand-in-hand with the accompanying increase in model size and complexity.

The most recent ``\ac{GenAI} era'' brought further advances, primarily with the introduction of \acp{DDPM} and other generative methods~\citep{Ho20-DDP}.
Moreover, the rapid explosion in \acp{LLM} and, increasingly, \acp{MFM} (i.\,e., models which can jointly handle multiple modalities, usually including language) opened up new avenues in the \emph{controllability} of \ac{ESS}~\citep{Triantafyllopoulos23-AOO}.
This is the state-of-play at the moment of writing.

The core idea behind modern-day, \acl{SGM} is to \emph{approximate} the generative distribution of the data they have been trained on; in the case of speech, this is the generative distribution that corresponds to natural speech.
Once this is done, these approximation models can be \emph{sampled} to synthesise realistic samples.
Nowadays, research primarily employs \acp{DNN} to learn this mapping from target text to waveform.
Much of the recent research on \acp{SGM} has been focused on improving the effectiveness of the models in order to more closely approximate the underlying distribution, as well as improve their controllability in order to make the generation process more malleable to the requirements of the user.
In the next subsections, we discuss how these models can be used to generate speech and vocal bursts, as well as how they can be explicitly controlled to procure the desired outcome.
We give a more detailed description of their mathematical underpinnings in \cref{ssec:sgm}.

\subsection{Expressions in speech}

\begin{figure}[t]
    \centering
    \includegraphics[width=\textwidth]{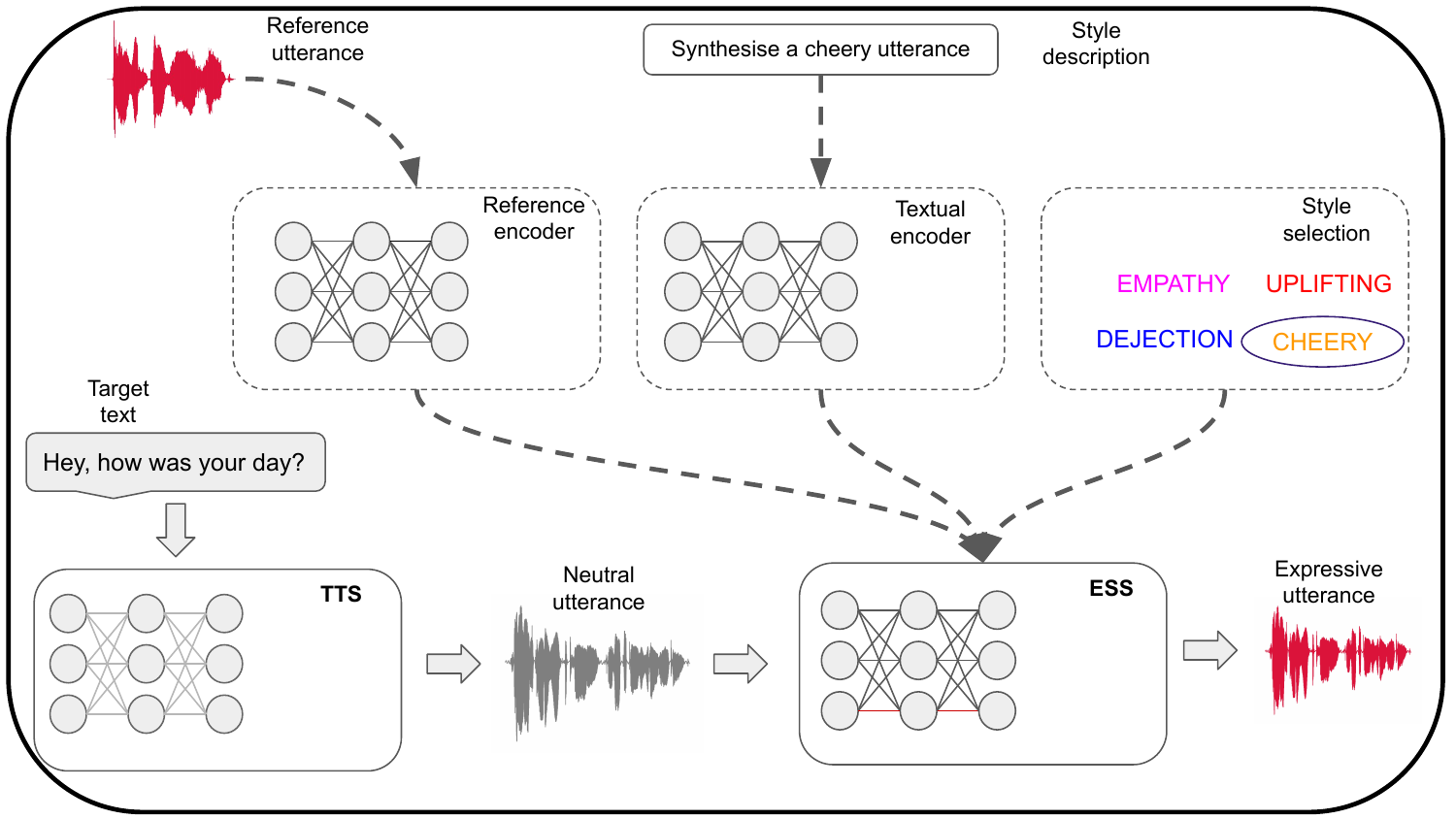}
    \caption{
    Overview of a typical ESS pipeline.
    An input text is first synthesised in a neutral style (gray) and then transformed to expressive speech (red) -- although these steps can also be integrated in an end-to-end model.
    The style is controlled either by a) a reference encoder which accepts as input a speech sample having the required style; b) a textual description in free text; c) a `tag' which allows the user to select from a fixed set of predefined styles.
    }
    \label{fig:ess}
\end{figure}

Synthesising expressive styles in speech entails the manipulation of those voice parameters which convey affective information: pitch, voice quality, rhythm, and pronunciation.
As we saw in \cref{sec:history}, in the early days of \ac{ESS}, these parameters were explicitly manipulated by rule-based systems.
With the recent rise of \acp{SGM}, the community has primarily focused on learning mappings from one expressive style (usually neutral) to another, with the models implicitly transforming those parameters as needed.

\cref{fig:ess} shows how this mapping can be achieved in practice:
Typically, the text that needs to be produced is transformed to a neutral utterance using \ac{TTS}; this text is determined based on a linguistic module, such as an \ac{LLM}; then, the appropriate style is applied to this utterance using a cascade voice conversion model.
In recent years, end-to-end models that start from text and directly output an expressive utterance are increasingly becoming the norm.
In either case, the expressive style is controlled as discussed in \cref{ssec:control} -- with a reference utterance, a linguistic prompt, a tag, or some combination of the above. 
The list of \ac{ESS} models is very long and beyond our scope here -- we refer to \citet{Triantafyllopoulos23-AOO} for a recent survey, although more and more models come out each year.

\subsection{Vocal bursts}
Non-verbal vocalisations, or ``vocal bursts'', also play an eminent role in expressing affect~\citep{Simon09-TVC}.
A timely exclamation can convey agreement, compassion, understanding, or support more easily than a verbal message, especially in the form of backchanelling during a real-time conversation~\citep{Hussain22-TSE, Cho22-AAA}.
When the demo for Google Assistant was unveiled, the crowd first erupted in cheers at the assistant's ``mm-hmm'' near the end of the video -- a startling depiction of the importance we place on vocal bursts~\citep{Triantafyllopoulos23-AOO}.
Until recently, however, their synthesis has received far less attention than their verbal counterparts.
This is quickly changing.
Notably, the Expressive Vocalisations (ExVo) series of workshops has called attention to their generation and provided the first challenge on synthesising vocal bursts, drawing increasing interest to this task~\citep{Baird22-TIE}.

In principle, the process for generating a vocal burst is similar to that of an expressive speech utterance and can be handled by a specialist \ac{SGM} trained explicitly for this task.
Perhaps even more so than verbal expressivity, the key challenge lies with knowing when to output such a vocalisation.
Timing is essential to transmit the appropriate message -- and this is where \emph{Stage II} \ac{ESS} becomes even more important.
We discuss this further in \cref{sec:complex}.

\subsection{Controllability}
\label{ssec:control}
The main hurdle to a successful, \ac{SGM}-based \emph{Stage I} \ac{ESS} system is achieving a satisfactory degree of \emph{controllability}~\citep{Triantafyllopoulos23-AOO}.
Controllability corresponds to conditioning the generation model (see \cref{ssec:sgm}) with additional information that \emph{guides} the generation process towards an output that matches certain requirements.

\textbf{One-hot encoding:} 
The standard form of conditioning relies on a constrained label space.
These labels encode the different states and traits that can be synthesised by the \ac{ESS} system.
The typical representation of those labels is a `one-hot' encoding, i.\,e., a 1D vector with dimensionality equal to the number of labels, populated with zeros everywhere except a single 1 in the element corresponding to the target class (some more advanced forms of encoding allow mixed classes; \cf \cref{sec:nuance}). 
In the simplest case, a different one-to-one model is trained and sampled for each state/trait combination; then, the label is simply used to select the appropriate model.
However, most recent works prefer to inject the label as additional information to the generating module (e.\,g., the decoder in an encoder-decoder architecture; see \citet{Rizos20-SFE, Triantafyllopoulos23-AOO}).

The major downside of one-hot encoding is that it only covers a restricted amount of expressive attributes that can be synthesised.
Moreover, given that it represents categories, it is only suitable for concepts that are categorical in nature.
For example, in the case of synthesising emotions, it is mostly used to synthesise categorical emotions, e.\,g., relying on Ekman's `big-6'~\citep{Ekman92-AAF}.
This is severely restricting the choices of \ac{ESS} creators, which is why the community is transitioning to the following two forms of conditioning.

\textbf{Audio prompts:} A more natural form of conditioning relies on (short) speech snippets that are uttered in the desired style~\citep{Wang17-TTE, Shen18-NTS, Skerry18-TEP}.
These short snippets are given as inputs along with an input text sequence or audio sample (synthesised in neutral voice) -- or both.
\ac{ESS} models that are controlled via audio prompts feature an additional prompt encoder, which maps the input audio to a set of embeddings that only encode the required style\footnote{Usually, this is ensured by supplementary training losses~\citep{Zhou22-EIC}.}.
The \ac{ESS} model then learns to map the input utterance to the target style specified by the additional prompt.
In the literature, this process is also known as \emph{reference encoding}~\citep{Wang17-TTE, Triantafyllopoulos23-AOO} or \emph{style transfer}~\citep{Jing19-NST}.

During training, these audio prompts are drawn from a pool of available data (oftentimes the same data the input and target utterances are drawn from).
During inference, they are instead given by the downstream user (though sometimes the user may select a style from some pool of references).

The main downside of auditory prompting is that the reference samples encompass a lot more information than the targeted expressive state~\citep{Triantafyllopoulos23-AOO}.
For example, they additionally include information about the prompt speaker's sex, age, ethnicity, or any other attribute that is encoded in the speech signal.
Moreover, as mentioned in the introduction of this chapter, this form of conditioning primarily follows the paradigm of a \emph{directionless} mimicking of a particular expressive state, without linking the expression that is generated to an underlying state.
It is thus particularly challenging to \emph{disentangle} all the different components such that only the required style is propagated to the main synthesis network~\citep{Wang19-STU}.
Recent works have taken aim at this challenge by introducing complementary losses to emphasise this aspect during training~\citep{Zhou22-EIC}, but despite their success, disentanglement remains an open problem.

\textbf{Linguistic prompts:} Lately, and especially with the recent rise of \acp{LLM}, it has become possible to condition \ac{ESS} models using linguistic prompts~\citep{Guo23-PCT, Leng23-P2D, Shimizu24-PCS}.
This is arguably the most natural form of controlling the models, as it makes it more intuitive -- and reproducible -- for downstream users.
The general layout is the same: the user gives an input prompt (``Generate a pleasing voice''), which is passed to a text encoder to generate embeddings that are then propagated to the synthesis module and the whole system is trained end-to-end (though some components might be frozen).

The intuition behind this form of conditioning is that the text encoder encapsulates knowledge about how a particular expression sounds.
In the case of \acp{LLM}, this knowledge is incorporated through its pretraining on very large corpora and uncovered via prompting or finetuning.
The downside is that the model may inherit the biases which accompany the text encoder(as is always the case using transfer learning; see also \citet{Bommasani21-OTO}).

\section{Stage II: Synthesising complex behaviours}
\label{sec:complex}
We noted in \cref{sec:taxonomy} how \emph{temporality} is a vital aspect of \ac{ESS}.
\cref{sec:primitives} discussed how \ac{GenAI} models can be used to synthesise a set of behavioural primitives -- simple affective states that can be understood within a singular utterance (or vocal burst).
The particular primitives that can be synthesised constitute the set of \emph{affordances} made available by a \emph{Stage I} \ac{ESS} system.
We now turn to how an \ac{AI} conversational agent can utilise these affordances to advance its \emph{Stage II} capabilities.

\begin{figure}[t]
    \centering
    \includegraphics[width=\textwidth]{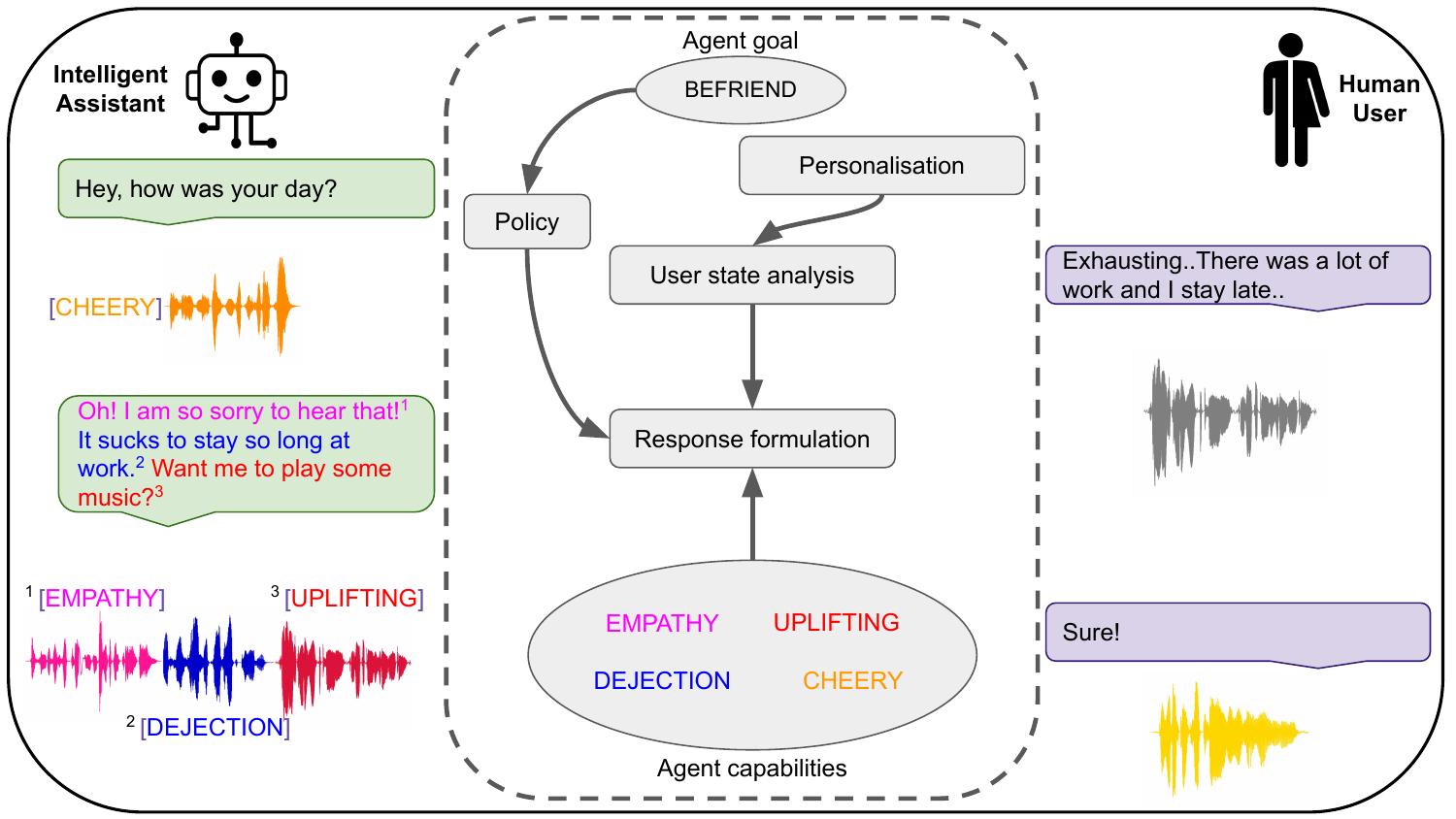}
   \caption{
   Overview of an \emph{Stage II} \ac{ESS} workflow, where an intelligent assistant pursues its overall goal of befriending a user, a goal which in turn guides each conversation.
   The middle panel shows the inner workings of the agent, while the side panels show the outcome of the conversation.
   The agent monitors the user's affective state and adjusts its responses accordingly, picking from an array of available expressive styles.
   We note that the styles available to the agent are not necessarily as interpretable as the ones we outline here; rather, we actually expect self-learning agents to develop their own internalised concepts which perhaps remain opaque to humans (see text for a more detailed discussion).
   }
   \label{fig:stage2:overview}
\end{figure}

\begin{figure}[t]
    \centering
    \includegraphics[width=\textwidth]{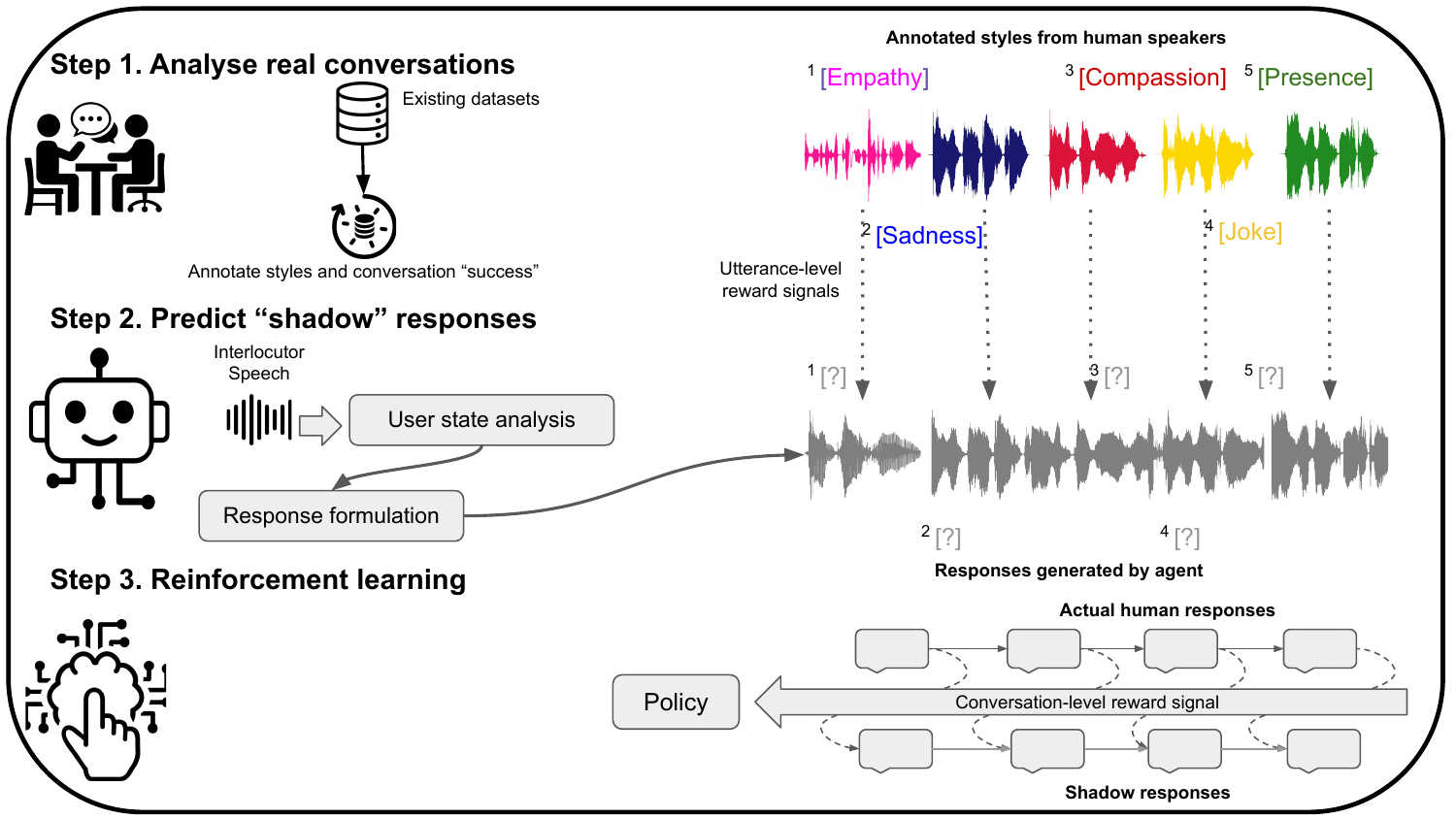}
   \caption{
   Blueprint for training a \emph{Stage II} \ac{ESS} pipeline.
   Training is first bootstrapped using available human-to-human conversations, with the agent rewarded for matching the next response by one of the two interlocutors.
   This initial training can be succeeded by a second reinforcement learning stage where the agent is fine-tuned on human-to-machine conversations, generated either \emph{on-policy} (i.\,e., by the agent being currently trained, perhaps in online fashion) or \emph{off-policy} (i.\,e., relying on prerecorded conversations).
   }
   \label{fig:stage2:training}
\end{figure}

\subsection{Learnt expressive policies}

A conceptual example for how a \emph{Stage II} \ac{ESS} agent operates is shown in \cref{fig:stage2:overview}.
It illustrates how an intelligent assistant with the overall goal to `befriend' an individual may utilise different expressive styles to achieve its goal depending on the context of an interaction.
It may begin with a [CHEERY] message, then understand that the user is in a rather dejected mood, and choose to respond with [EMPATHY] and its own [DEJECTION], before proceeding with another attempt for an [UPLIFTING] utterance.
The choice of styles and their ordering is set by the agent's \emph{policy}, which takes into account the present interaction and the user's overall preferences.

The expressive primitives available by \emph{Stage I} \ac{ESS} can thus be considered as a set of available \emph{actions}\footnote{These actions are intentionally reminiscent of acts in speech act theory~\citep{Austin75-HTD}. However, they are not entirely the same concept. While their aim is to help `achieve' the agent's goal(s), they do not have a direct mapping to the performative context of some speech acts.} that an agent can choose from at each conversational turn (or even in-between, in the case of backchanneling).
These actions must be combined over several turns to achieve a particular \emph{goal}\footnote{In general, they might also have to be combined over several interactions with the same users, for example in order to communicate some long-term state like political stance or ideology.}.
This high-level goal can be achieved by utilising an appropriate chain of actions.

Importantly, the \emph{policy} for selecting the appropriate set of actions can be either hardcoded or learnt.
In the latter case, our conceptualisation is amenable to a standard \ac{RL} framing~\citep{Sutton18-RLA}.
\cref{fig:stage2:training} shows a blueprint for how a policy can be learnt from conversational interactions:
An initial bootstrapping phase allows learning to generate appropriate responses from observing human-human conversations; in that phase, the agent is tasked with generating ``shadow'' responses for one or both of the interlocutors, and attempts to match the actual response of a human.
The underlying assumption here is that humans more often than not pick the most suitable response in a conversation.
This assumption can be relaxed by annotating some conversations with respect to how `successful' they were (although we still expect a large benefit from pre-training on unlabelled large data before using annotations).
On a second step, the agent is used for actual human-machine conversations, where it receives a positive reward when it achieves its goals.
There, the agent actually partakes in a conversation, selects the most suitable action combination using its present policy, receives its reward in the form of feedback, and updates its policy accordingly.

Similar to contemporary \ac{RL} systems, the process can involve human interlocutors (i.\,e., a \ac{RLHF} setup as proposed in \citet{Griffith13-PSI}) or be bootstrapped with self-play using artificial agents~\citep{Silver18-AGR} -- both strategies which have proven successful in other \ac{RL} problems.
Moreover, the hierarchy of system goals can be extended to more levels than two.
In our example, [EMPATHY] may be a sub-goal of [BEFRIEND], which in turn might be a sub-goal of [SUPPORT], or even, on a potentially more sinister turn, of [INFLUENCE] and [DECEIVE].
This framework thus allows for an extension of the affordances that an agent can employ -- or even autonomously acquire throughout its life-cycle.
We note that we have selected these styles for illustration purposes only; in fact, we expect \ac{ESS} models to rely on internalised constructs that remain, to a smaller or greater extent, uninterpretable to humans (especially when incorporated in the inner space of foundation models; \cf \cref{sec:foundation}).

Intriguingly, \ac{RL} training may further lead \ac{ESS} models to uncover entirely novel forms of expression that are not (presently) used by humans.
We note that the emergence of new expressive styles is common for humans -- and is further impacted by modern media~\citep{Androutsopoulos14-MAS}.
In principle, there is nothing preventing artificial \ac{ESS} models to introduce such styles, which humans may then choose to mimic, either explicitly or implicitly (i.\,e., simply owning to the popularity of those styles in social media and beyond).

\subsection{Mixed-state synthesis}
\label{sec:nuance}

Oftentimes, the state that an agent needs to express is \emph{mixed} and comprises multiple simpler states that need to be synthesised jointly.
Stress is a good example.
\citet[p. 35]{Lazarus99-SAE} argued that:
``When there is stress there are emotions... when there are emotions, even positively toned ones, there is often stress too...''
Another one, perhaps more pertinent to affective agents, is \emph{compassion}~\citep{Goetz10-CAE}.

Some research efforts have targeted the synthesis of mixed states~\citep{Zhou22-SSW}.
Typically, these try to \emph{interpolate} between two or more `clear-cut' elements, thus resulting in a new construct that lands somewhere in between.
For instance, happiness and sadness can be interpolated to obtain a \emph{bittersweet} state.

Technically, this effect can be achieved by actually interpolating between the embedding spaces characterising the two initial states; assuming that this space is sufficiently well-behaved, a mathematical interpolation (e.\,g., averaging or taking the geodesic mean) will result in a point in that space that maps to an `in-between' state.
This is congruent with the \emph{semantic space theory} recently proposed by \citet{Cowen21-SST}, which postulates that emotions exist on a well-behaved manifold, whose traversal yields smooth transitions between the different emotions.

However, there exists another side to synthesising mixed-states.
Previous work has focused on expressing a single new state which characterises an entire utterance -- this is equivalent to our \emph{Stage I} \ac{ESS} capabilities discussed previously.
The corresponding \emph{Stage II} implementation would require the interlacing of multiple states in a longer segment comprising multiple utterances, some of one state, some of the other -- and some in-between.
Crucially, planning the trajectory between successive states to achieve the required effect, as well as managing the corresponding transitions, requires the higher-level capabilities of a more advanced policy of the kind envisioned above.
This still remains open-ground for \ac{ESS} systems.

\subsection{Personalised expressive speech synthesis}
\label{ssec:personalisation}
Expressivity exists in the ears of the beholder.
How a speaker is perceived depends on the background and current affective state of the listener.
Previous research suggests that personal effects mediate both the perception of expressive primitives and more complex behaviour.
For example, different age, gender, and culture groups have been found to perceive emotions differently~\citep{Dang10-COE, Ben19-ADI, Zhao19-DTL}, while the relatively low to moderate inter-annotator agreements found in several contemporary \ac{SER} datasets demonstrates how individualised the perception of affect may be (e.\,g., see \citet{Lotfian17-BNE}).
Consequently, this means that \ac{AI} conversational agents must learn to adapt in-context to their interlocutor, a feat that is being pursued for \acp{LLM}~\citep{Kirk24-TBR}.

To achieve this, it is necessary for \ac{AI} agents to actively monitor their interlocutors during a conversation.
On a first level, they may try to identify their demographics; this helps frame their interlocutor as belonging to particular groups with known preferences.
On a deeper layer, the agents must identify how their interlocutors perceive expressivity.
This can be a achieved by a trial-and-error process, where the agent makes attempts to express particular states and gauges the response they elicit.
Essentially, this fits into the \ac{RL} framework outlined in \cref{sec:complex}, where the agent first selects an appropriate \emph{action} given its current \emph{policy} and overarching \emph{goal}, and subsequently updates that policy given the \emph{reward} received by the `environment' (i.\,e, the difference between the user state that the agent aimed for and the one it actually elicited).

Finally, we note that this interpersonal adaptation additionally subsumes \emph{entrainment}~\citep{Brennan96-CPA, Amiriparian19-SII}.
Entrainment is the process of convergence that happens between two conversational partners, a prerequisite for a successful and enjoyable conversation.
It manifests as a gradual adaptation to the linguistic structure and the paralinguistic expressions of the other -- and generally happens from both partners.
Conversational agents must therefore include entrainment in their affordances.
This requires an analysis (implicit or explicit) of their interlocutor's manner of speaking.
This analysis can be coupled with the state and trait analysis mentioned above, and can broadly cover more granular paralinguistic markers such as pitch, tempo, and even timbre, or linguistic behaviour ranging from word-use to deeper grammatical structure.

Overall, we consider personalisation to be an exciting new frontier for \ac{ESS} (and \acp{LLM}, see \citet{Kirk24-TBR}), given that most existing systems lack a `feedback mechanism' that allows them to adapt to each new user.
Prior research has been largely focused on obtaining good, `universal' expressivity, but with that goal now closer to sight, it might be time to switch to a more modular, malleable, and adaptive approach.

\section{Foundation models and emergence}
\label{sec:foundation}
The introduction of \acp{FM}, and especially generative \acp{LAM}, has paved the way for a novel paradigm of synthesis, especially as it pertains to controllability.
Specifically, while \acp{LAM} (so far) follow the same basic operating principles as traditional \acp{SGM}, their scale and amount of data they have consumed in training gives rise to \emph{emergent} properties -- properties that have not been explicitly trained for but are uncovered using appropriate prompting~\citep{Wei22-EAO}.

This phenomenon was first observed for \acp{LLM}, which were shown capable of performing tasks that were not part of their training simply by providing them with appropriate prompts~\citep{Wei22-EAO}.
In similar fashion, \acp{LAM} can be prompted to synthesise styles that were not part of their training.
In practice, `all' a \ac{LAM} does is encapsulate all the individual steps shown in \cref{fig:ess} in one singular architecture.
Such models accept multimodal inputs in a more modular way; parts of them correspond to the output text that must be generated, and parts pertain to the style that needs to be synthesised.
Crucially, inputs can also incorporate \emph{Stage II} capabilities; for instance, the overall goal of the system or information about the user may be given as part of the prompt~\citep{Kirk24-TBR}.

For example, an input prompt might be: ``You are an intelligent assistant aiming to befriend the user. The user is a male computer science student that just returned home from the lab. Generate an uplifting message to start a conversation.''
This modularity enables \acp{LAM} to benefit from the compositionality of language -- rather than trained to synthesise specific styles explicitly, they learn to map longer text inputs which consolidate information about style, intent, and context into an output utterance. 
This allows a more flexible interface that can scale to novel situations by tapping into the world knowledge of the model.
Namely, while a traditional generative model would need to be trained to generate happy, cheerful, or compassionate speech explicitly, a \ac{LAM} only needs to exploit its understanding of each term (as well as its understanding of how to synthesise expressive speech) and achieve the task without having been trained for every style explicitly (though it will, of course, need to be trained for some of them).
This feat alone unlocks a tremendous potential for scalability.

Naturally, given the success of \acp{LLM} in chatting but also longer-term planning, it will also be straightforward to include \ac{ESS} along with the text-generating and dialogue management capabilities of existing models, and even let the model pick the suitable style on its own, thus resulting in a truly end-to-end artificial conversational agent.
This means that the paradigm of foundation models has the potential to resolve a lot of the issues that are still open for \ac{ESS}.
Even though we are still in the early days of their development, the recent experience with \acp{LLM} and vision \acp{FM} points to a (near) future where \acp{LAM} become the norm~\citep{Bommasani21-OTO}.

Consequently, this state of affairs raises the same considerations as for \acp{FM}~\citep{Bommasani21-OTO}, namely, regarding \emph{fairness} and the \emph{representation} of different socio-demographic groups in the data;
the \emph{alignment} of those models with established ethical values;
the models' \emph{computational cost}, both in terms of harm done to the environment and with respect to the limited access to that technology that the increase in computational complexity entails;
the lack of \emph{interpretability};
the appearance of \emph{hallucinations}, with models failing to follow instructions but nevertheless producing outputs which sound plausible;
and, finally, the potential that any advances introduced by \acp{FM} can be subverted by bad actors for nefarious purposes\footnote{Though this is true for any technology, we decided to mention it explicitly given phrased concerns by regulators around the world, as seen, for example, with the EU AI Act which directly mentions foundation models~\citep{AIAct}.}.
This is the topic of our last section.

\section{Societal implications of advancing ESS systems}
\label{sec:metaverse}
In this section, we discuss the societal implications that accompany the advancement of \ac{ESS} research.
This pertains both to its current state, but also to the expected advances we outlined in previous sections.

\subsection{Persuasion and manipulation}

One of the most obvious downsides to \ac{ESS} is its potential for misuse.
Crafting more expressive artificial voices unlocks the possibility to scale up misinformation, unwarranted persuasion, manipulation, or outright fraud.
The use of voice cloning -- a sister field of \ac{ESS} where the goal is to simulate the identity of a particular human speaker -- is raising increasing concerns.
This technology can be misused to impersonate family members or persons of authority in order to manipulate the victim, a process that is already causing pressing societal problems.
However, \ac{ESS} will allow fraudsters to progress even beyond that by leveraging more advanced intelligent agents to persuade their subjects.

In the last few years, claims have emerged that text generated using commercial \acp{LLM} can be more persuasive than human-generated text~\citep{Hackenburg23-ETP, Durmus24-MTP, Salvi24-OTC}.
While these findings are still preliminary, they nevertheless showcase the feasibility of using computer-generated speech for persuasion.
We expect \ac{ESS} to further advance this potential, as integrating paralinguistic cues can increase the effectiveness of the message~\citep{Van20-HTV}.
Notably, personalisation (in the sense of adapting to user demographics or prior opinions) led to performance improvements, a theme we also highlighted in \cref{ssec:personalisation}.

\subsection{A metaverse of superhuman influencers}

Beyond fraudulent or criminal behaviour, \ac{ESS} may have negative effects even under lawful usage.
Specifically, we expect that as the more advanced models we described in previous sections become increasingly available, they will be used by a number of actors to generate or manipulate digital content in order to make it more \emph{appealing}.
This means that the voices we encounter in digital spheres will come to be increasingly enhanced, or even fully generated, by \ac{AI}.
As such, society might soon find itself in a \emph{metaverse} populated with superhuman `influencers' -- agents, human or artificial, who possess above-average charisma and expressivity.
This calls into question the changes this might impart on expressivity and language itself.
This much broader field of study falls under the premises of \emph{sociolinguistics}, which studies the interaction between social and linguistic change, oftentimes in the landscape formed by modern media~\citep{Androutsopoulos14-MAS} (Expressivity and the Media, this volume).
In the following paragraphs, we highlight some particular repercussions of \ac{ESS} being deployed in the real-world.

The first frontier is attention.
Commercials are already mired in what has been labelled a ``loudness war''~\citep{Moore03-WAC}.
Yet the impact of such simplistic forms of manipulation that rely on a single cue (loudness) to draw our attention pales in comparison to the potential wave of information streams augmented with the use of advanced \ac{ESS} systems.
Given that attention has become a commodity in today's ``attention economy''~\citep{Davenport01-TAE}, this could inspire renewed competition between commercials, news sites, and anyone else vying for our focused engagement in the digital sphere.

Especially for digital media, there are not many studies on the interplay between voice expressivity and social media dynamics (like engagement or outreach).
We can, however, draw some insights from similar studies on visual aesthetics.
For example, fashion brands that opt for a more expressive style in their posts (vibrant colours, modern design, energetic) have a bigger outreach than others who prefer more classic aesthetics (orderly and clear design; \citet{Lavie04-ADO, Kusumasondjaja20-ETR}).
Similarly, specific speech attributes can lead to improved visibility in social media, which in turn creates incentives that `select' for these attributes to be more widely used.

This becomes even more pertinent when considering overall social media use.
Overexposure to social networking sites, especially when consumption is focused on perusing profiles heavy on visual content, has been linked to reduced self-esteem and negative self-evaluation~\citep{Lee14-HDP, Vogel14-SCS, Cohen17-TRB}.
While previous studies have focused on the visual component of social media, they have largely ignored the fact that they are also rife with spoken content.
Assuming the ubiquitous presence of \ac{ESS} in the near future, and especially models optimised for human voice enhancement, we can expect that these models will be used to improve the outreach and appeal of social media content, similar to how facial `filters' are used today.
This raises the question of how social media users will react to a virtual world filled with oversaturated and overexpressive voices.
Authentically charismatic speakers can use their voice to stand out of the crowd, but this ability may soon become available for everyone by simply using an off-the-shelf \ac{ESS} model.
On the one hand, this will level the playing field for people competing for our attention.
On the other hand, it will lead to an overexposure to charismatic speakers, which will inadvertently feed into changes of our aesthetics.
Whether this will lead to a desensitisation effect, where people learn to ignore paralinguistic styles previously associated with charisma, or open up our senses to previously unappreciated modes of expression remains to be seen.
In any case, we expect \ac{ESS} to become a staple in the toolkit of professional influencers\footnote{TikTok, for example, features ``voice effects''~\citep{TikTok}.}.


\subsection{Aligned artificial expressivity}
Given some of the negative social implications that \ac{ESS} might cause, it is important to ensure that all \ac{ESS} systems remain \emph{aligned} with societal values -- i.\,e., making \ac{ESS} ``friendly''~\citep{Yudkowsky01-CFA}.
Naturally, we expect this process to involve stringent regulatory guidelines, such as banning the use of unsolicited deepfakes, and the accompanying effort to enforce them, which are beyond our scope here.
Instead, we consider how model development can prevent even lawful uses of the technology from going awry.

We expect that ethical and regulatory guidelines will have to be `baked into' a model's behaviour during training, especially with regards to its \emph{Stage II} capabilities.
Inspiration for this can be drawn from the recent advances in \emph{physics-informed neural networks}~\citep{Raissi19-PNN}, which solve physics-related problems (e.\,g., material design) using \acp{DNN} but explicitly guide these \acp{DNN} to generate outputs that conform to physical laws.
Similarly, we can envision \ac{ESS} systems whose outputs conform to judicial laws and social ethics.
For example, an \ac{ESS} system that is deployed for online marketing could refrain from using persuasion techniques on particularly vulnerable users, such as children.
Training \ac{ESS} models -- and especially the most recent foundation models -- to conform to those norms is an area of active research.
In terms of training, it boils down to additional constraints that need to be satisfied.
These could be implemented by extending the loss function of a model or optimising its parameters in a constrained optimisation paradigm -- similar to disentanglement (\cref{ssec:sgm}).


On top of that, auditing whether \ac{ESS} systems adhere to all guidelines in practice is a much more challenging endeavour.
Even assuming that models are publicly accessible (e.\,g., through APIs available for research purposes), testing them rigorously, and periodically, to cover new releases, remains an open issue.
In the most straightforward case, this would involve the use of human auditors -- test users who interact with \ac{ESS} systems and rate their abilities.
However, this process does not scale well in practice due to the sheer number of vendors and open-source models that are even now available.
Moreover, humans will inevitably bump into the measurement issues outlined in \citet{Triantafyllopoulos23-AOO}; lay users, in particular, might struggle with advanced \ac{ESS} models that use nuanced strategies for manipulation.
In the most extreme case, we can envision self-learning \ac{ESS} systems developing capabilities that enable them to circumvent testing, a case of a so-called ``Runaway AI''~\citep{Guihot17-NRI}. 

For all those reasons, we expect machine-based auditing to become increasingly more relevant as it offers better scalability and reproducibility.
Such models are being developed to identify spoofing attempts~\citep{Liu23-ASV} -- \ac{AI}-generated speech that is intended for malicious purposes such as identity theft.
While these models are failing to capture all speech samples generated by contemporary \ac{TTS} systems, they do work to some extent and are a useful tool in mitigating threats resulting from unlawful use of the technology (e.\,g., identity theft).
A similar effort is required to monitor lawful but ethically dubious use of \ac{ESS} technology.
While challenging, we expect this pursuit to be fruitful and a critical step in ensuring the fair and ethical development of artificial expressivity.

\section{Summary \& Conclusion}
\label{sec:conclusion}
We have presented an overview of the fundamental blocks required to build \emph{\acl{ESS}} systems, starting from nowadays standard statistical generative models and reaching to recently-introduced foundation models.
We have also discussed open risks and highlighted areas which we expect are ripe for innovation.
In summary, we see a consolidation of \ac{ESS} into the broader move towards foundation models which encapsulate multiple, often wildly disparate, capabilities, as well as the emergence of longer-term planning and in-context synthesis (what we termed \emph{Stage II} capabilities).
This is an exciting time for \ac{ESS} research, as the last handful of years have seen tremendous progress in fidelity and controllability for synthesising expressive primitives -- elementary styles that can form the basic building blocks for more complex behaviour.
We anticipate that the landscape of \ac{ESS} over the next decade will be largely defined by the pursuit of methods that can combine these primitives and translate them into more nuanced states, as well as a drive to ensure that models remain tethered to social and ethical norms.

\appendix

\section{Statistical generative models}
\label{ssec:sgm}

In this appendix section, we discuss how \ac{ESS} can be used to synthesise what we term expressive \emph{primitives} -- transient behaviours which are completely encompassed within a single episode.
These primitives include very short states which dominate behaviour for a fixed period of time, like emotions, and immutable traits which are omnipresent, such as gender or age~\citep{Schuller14-CPE}.
As such, they can be portrayed within the confines of a single utterance.
This can be done by manipulating the paralinguistic structure of the utterance, as well as by introducing vocal bursts as short interjections that convey a particular affect.
Both tasks are achieved by following the principles outlined below -- one needs to train a model on data that encompass the targeted expressive behaviours.

As discussed in \cref{sec:history}, contemporary \ac{ESS} is statistical in nature, falling under the auspices of \ac{ML}, and specifically \acp{SGM}.
\acp{SGM} constitute a model of the underlying data generation process; as such, they allow sampling from that process to generate new content.
Traditionally, the generation process to be modelled was that of generating speech from text (i.\,e., \ac{TTS}) and converting it to some expressive state.
Early \acp{SGM} were \emph{specialists}, focused exclusively on particular mappings, namely, the ones defined by the data and tasks they were trained on.
We describe their inner workings in the following subsections.

In the present subsection, we begin with a quick overview of the mathematical underpinnings of \acp{SGM} followed by a discussion of the most crucial components that are needed to build \ac{ESS} systems.

\subsection{Preliminaries}
Broadly, \acp{SGM} can be seen as a category of models $f_{\pmb{\theta}}$ that aim to capture the data generating distribution:
\begin{equation}
    f_{\pmb{\theta}}: \mathbb{R}^N \rightarrow [0, 1],
\end{equation}
with $N$ being the dimensionality of the output signal\footnote{
In theory, this can reach up to $\infty$ for speech signals.
In practice, though, it is often bounded to a few seconds.
}.
This $f$ is usually trained to approximate the \emph{true} data distribution process:
\begin{equation}
    p(x_1, ..., x_N).
\end{equation}
Usually, the latter is expected to be \emph{conditioned} by some additional information $y$, in which case it takes the form:
\begin{equation}
    p(x_1, ..., x_N| y),
\end{equation}
with $y$ now taking arbitrary values (e.\,g., a class label denoted as integer or even text; see below).

In order to ensure that $f(\cdot)$ is a proper probability distribution, it is often thought of as the \emph{normalised} form of an \emph{unnormalised} energy distribution $E$\footnote{The term ``energy'' is used because this particular model describes the distribution of particles according to the Boltzmann-Gibbs distribution in statistical mechanics.}:
\begin{equation}
    f_{\pmb{\theta}}(\pmb{x}) = \frac{e^{-\beta E(\pmb{x})}}{\int_{\pmb{c} \in \mathcal{D}}e^{-\beta E(\pmb{c})}},
    \label{eq:energy}
\end{equation}
with $\pmb{x}=(x_1, ..., x_N)$, $\pmb{c}=(c_1, ..., c_N)$, $\mathcal{D}$ being the \emph{set of all possible data points}, and $\beta$ a normalisation (temperature) parameter which we will henceforth ignore.
We note that the major bottleneck in computing $f$ is the presence of the integral of $\mathcal{D}$ in the denominator; in the general case, this must be evaluated over all data points (i.\,e., the space of all possible speech utterances in our case).
This denominator is often referred to as the \emph{partition function} $Z_{\pmb{\theta}}$ and is considered intractable for most practical applications.
We return to this point when we discuss how these models are actually trained in \cref{ssec:training}.

Broadly, we distinguish between two main forms of \ac{ESS} systems, depending on their input-output schemas:
\begin{enumerate}
    \item \emph{Expressive TTS:} these ``end-to-end'' models generate expressive speech directly from text; thus, they directly create an utterance in expressive style.
    \item \emph{Expressive voice conversion:} these ``cascade'' models manipulate an input speech signal to change its expressive style; usually, they are combined with a `simple' \ac{TTS} frontend that creates a speech utterance in neutral style, which is then transformed by the \ac{ESS} model.
\end{enumerate}
An overview of both and their differences can be found in \citet{Triantafyllopoulos23-AOO}.

There are three main challenges associated with $f_{\pmb{\theta}}$:
\begin{enumerate}
    \item \emph{Training} it to become a good approximation of $p(\cdot)$;
    \item Being able to \emph{sample} efficiently from it;
    \item Achieving good levels of \emph{control} for the different values of $y$.
\end{enumerate}
We discuss each of them in the subsections that follow.

\subsection{Training}
\label{ssec:training}

\acp{SGM} are trained on (large) corpora of speech; in the case of expressive \ac{TTS} they are trained to output speech from text (either graphemes or phonemes); in the case of expressive voice conversion, they are instead trained to map speech to speech.
We note that our goal during training is to estimate the normalised energy function $f_{\pmb{\theta}}(\cdot)$ from \cref{eq:energy}.
This is achieved by using a training set $\mathcal{S}$ and computing the function $f_{\pmb{\theta}}(\cdot)$ that maximises the \emph{likelihood} over $\mathcal{S}$; in layman's term, this optimal $f^*_{\pmb{\theta}}(\cdot)$ is the model which best captures the variability over the observed data -- the most ``likely'' model given the evidence.
As the standard algorithm used for training (especially for neural networks) is (stochastic) gradient descent, in practice, we \emph{minimise} the negative likelihood -- and a logarithm is often taken to remove the exponent.
Thus, we end up with the following negative loss-likelihood loss function $\mathcal{L}_{NLL}$:
\begin{align*}
		\mathcal{L}_{\text{NLL}} &= -\mathbb{E}_{\mathcal{S}}[L(\pmb{x})] \\
            &= -\mathbb{E}_{\mathcal{S}}[log(f_{\pmb{\theta}}(\pmb{x}))] \\
		&=  -\mathbb{E}_{\mathcal{S}}[log(\frac{1}{Z_{\pmb{\theta}}} e^{-E_{\pmb{\theta}}(\pmb{x})})]\\
		&=  \mathbb{E}_{\mathcal{S}}[E_{\pmb{\theta}}(\pmb{x})] +\mathbb{E}_{\mathcal{S}}[log({Z_{\pmb{\theta}}})]\\
		 &=  \mathbb{E}_{\mathcal{S}}[E_{\pmb{\theta}}(\pmb{x})] +log({Z_{\pmb{\theta}}}) \\
          &=  \mathbb{E}_{\mathcal{S}}[E_{\pmb{\theta}}(\pmb{x})] + log(\int_{\pmb{c} \in \mathcal{D}}e^{-E_{\pmb{\theta}}(\pmb{c})})\\
          &=  \int_{\pmb{x} \in \mathcal{S}}E_{\pmb{\theta}}(\pmb{x})] - \int_{\pmb{c} \in \mathcal{D}}E_{\pmb{\theta}}(\pmb{c}).
\end{align*}
Note that the first term above (often referred to as the ``positive phase'') is computed over the training set $\mathcal{S}$; the latter (the ``negative phase''), is instead computed over the entire true distribution of data $\mathcal{D}$.
The positive phase increases the likelihood of the observed data; the negative phase in turn grounds that likelihood by keeping it limited over the entire space of possible data.
Importantly, during training with gradient descent, both integrals are approximated using a sum (over the finite set of observed data); this is also the process used by the popular stochastic gradient descent algorithm and its variants.
However, in each iteration, one must also evaluate the negative phase over $\mathcal{D}$.
There are two issues with this:
\begin{enumerate}
    \item The computational overhead of always evaluating the value of the energy function over $\mathcal{D}$ is intractable.
    \item More importantly, it is almost impossible to observe this $\mathcal{D}$ in practice; not only does it include all possible observable data points (e.\,g., all possible speech utterances that will ever be uttered in the entire history of humanity in our case), but in the strict sense, it also includes all `garbage' sounds that fit into the embedding space defined by $\pmb{x}$; technically, even though these sounds will have a very low probability, they still need to be evaluated.
\end{enumerate}

The above two bottlenecks make it very hard to identify a suitable $\mathcal{D}$ to integrate over.
All modern variants of \acp{SGM} are explicitly aimed at overcoming this hurdle:
\acp{VAE} circumvent the need to approximate the partition function by optimising instead a lower bound, the so-called evidence lower bound (ELBO)~\citep{Doersch16-TOV}.
Contrastive methods increase the likelihood on observed data and decrease it on fake data~\citep{Hinton02-TPO}; the difference between those two likelihoods eliminates the necessity to compute the partition function.
\acp{DDPM} rely on score matching~\citep{Ho20-DDP, Hyvarinen05-EON}, whereby the dependence on $Z_{\pmb{\theta}}$ is lifted by substituting the estimated likelihood with its derivative.
Our focus here, however, is not on thoroughly reviewing these (and other) methods, so, we instead refer the reader to relevant surveys~\citep{Tan21-NSS, Cao04-ASO}.
For our purposes, it is important to note that \acp{DDPM}~\citep{Ho20-DDP} have emerged as the most recent class of methods with impressive generative results, and are nowadays the go-to method for most \ac{GenAI} applications, including \ac{ESS}~\citep{Popov21-GAD, Huang22-PPF, Prabhu24-EDB}, at least in terms of offline generation.

\subsection{Sampling}
After successfully training an approximation of $p(\cdot)$, it becomes necessary to sample from it during inference.
This is also not a trivial problem, especially because the typical forms of $f_{\pmb{\theta}}$ are complex and make sampling complicated.
A key challenge arises from the fact that $N$ is high-dimensional -- particularly for \ac{ESS}.
For example, assuming we aim to generate a 1-second sample at 16\,kHz, then a model needs to procure 16,000 samples.
\cref{alg:gibbs} shows how this sampling can be achieved with a simplified\footnote{In practice, the initial step is not Gaussian but is usually derived from the training data.} version of a traditional algorithm, namely, Gibbs sampling, an instance of a \ac{MCMC} method~\citep{Gelfand00-GSA}.
Gibbs sampling relies on the iterative sampling of all variables by using the conditional distribution of that variable over all others.
The iteration stops once all variables have converged.
It is evident that performing this procedure for 16,000 samples -- let alone for longer sequences -- easily becomes computationally prohibitive depending on the structure of $f_{\pmb{\theta}}$.

\begin{algorithm}[t]
\caption{
Example of a typical sampling algorithm:
Gibbs sampling with Gaussian initialisation.
}\label{alg:gibbs}
\begin{algorithmic}
\State $x^0_i \gets x \sim \mathcal{N}(0, 1), \forall i \in {0, ..., N}$
\While{$x^t_i - x^{t-1}_i > \epsilon  \forall i \in {0, ..., N}$}
\ForAll{$i \in {0,...,N}$}
    \State $x^t_i \gets x \sim f_{\pmb{\theta}}(x^{t-1}_i|x^{t-1}_1, ..., x^{t-1}_{i-1}, x^{t-1}_{i+1}, ..., x^{t-1}_N)f_{\pmb{\theta}}(x^{t-1}_1, ..., x^{t-1}_{i-1}, x^{t-1}_{i+1}, ..., x^{t-1}_N)$ 
\EndFor
\EndWhile
\end{algorithmic}
\end{algorithm}

To overcome this crucial challenge, the community has focused on \acp{SGM} that can be efficiently sampled.
Here, \acp{DDPM} suffer from an additional overhead imposed by iterating over the denoising distribution~\citep{Ho20-DDP, Song20-ITF} and are thus not suited for the real-time requirements of some \ac{ESS} applications (\cf \cref{sec:applications}).
While recent efforts have been targeted towards addressing this bottleneck~\citep{Song03-CMO}, the current state-of-the-art relies on slightly older methods, primarily autoregressive models~\citep{Tan21-NSS}.
	
\printbibliography

\end{document}